# Dynamically Computing Adversarial Perturbations for Recurrent Neural Networks

Shankar A. Deka, Dušan M. Stipanović , *Member, IEEE,* and Claire J. Tomlin, *Fellow, IEEE*

*Abstract*—Convolutional and recurrent neural networks have been widely employed to achieve state-of-the-art performance on classification tasks. However, it has also been noted that these networks can be manipulated adversarially with relative ease, by carefully crafted additive perturbations to the input. Though several experimentally established prior works exist on crafting and defending against attacks, it is also desirable to have theoretical guarantees on the existence of adversarial examples and robustness margins of the network to such examples. We provide both in this paper. We focus specifically on recurrent architectures and draw inspiration from dynamical systems theory to naturally cast this as a control problem, allowing us to dynamically compute adversarial perturbations at each timestep of the input sequence, thus resembling a feedback controller. Illustrative examples are provided to supplement the theoretical discussions.

*Index Terms*—Adversarial examples, Recurrent neural network, Control synthesis, Dynamical systems.

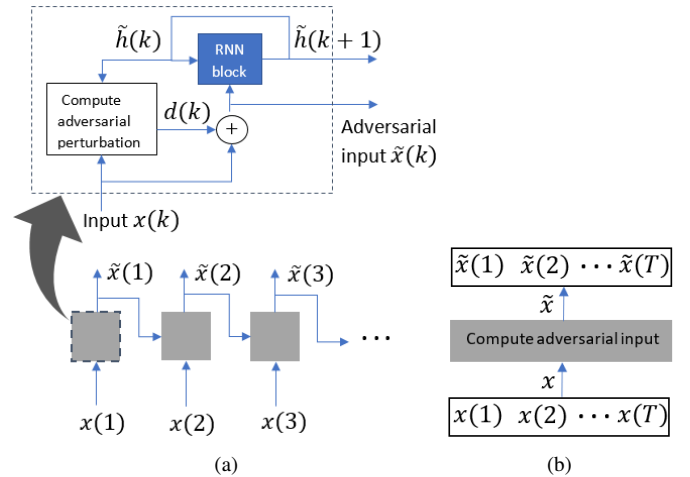

Fig. 1. (a) Proposed algorithm for RNNs dynamically computes the adversarial disturbance at each timestep enabling 'real-time' injection of adversarial perturbation (b) By contrast, a prior algorithm, such as FGSM, would require the entire input sequence collated together, before it can compute the perturbation.

## I. Introduction

ADVERSARIAL attacks on neural networks, and techniques to robustify against such attacks, have been a topic of continued interest in the machine learning communities. A wide variety of methodologies has been presented in the past, for crafting adversarial input disturbances. These broadly fall under one category or another based on the information available to the attacker regarding the neural network being attacked. For example, widely known methods like Fast Gradient Sign Method (FGSM) [1] or Carlini and Wagner (C&W) attack [2], Jacobian-based saliency map attack [3] to name a few, were developed considering a white-box attack model. Others like [4], [5], [6], [7] work for black-box attacks. 'Transfer attack' models like [8] do not assume knowledge of the model but require the training data. As far as the applications are concerned, the majority of prior works focus on computer vision [9], [10], [1], wherein given a feed-forward network (usually a convolutional neural network (CNN)) and an input image, an adversarial disturbance is crafted to fool the network on image classification tasks. A small few, like [11], have considered attacks on Recurrent Neural Networks (RNNs), although their algorithm requires unfolding the RNN first, thus resembling a feed-forward network in practice. Observations drawn from these prior empirically-supported algorithms and results have often led to conflicting conclusions. For instance, works such as [12], [13] speculate that adversarial examples are confined to tight "pockets" due to the highly nonlinear data manifold that the network represents, whereas [1] have argued that it is in fact the contrary, i.e., "models being too linear" is what gives rise to adversarial examples. This motivates the need for a more rigorous mathematical formulation of the adversarial perturbation problem, which would hopefully shed light on how the input perturbations impact the network output from the nominal input, and under what conditions adversarial perturbations are guaranteed to exist. From myriad experimentation in previous work, it appears that any given input can be perturbed adversarially using some simple gradient-based disturbance. A complementary line of work, on defense methodologies against adversarial inputs, has been considered by [14], [15], [16], with the simplest approach being augmentation of training data with adversarial examples.

In this paper, we consider white-box attacks specifically on RNN models and demonstrate how their recurrent nature can also be utilized to compute disturbances dynamically. This is the key difference from (very limited) prior works on adversarial attack algorithms applicable to RNNs. The contributions of our paper are as follows:

- We present an algorithm to *sequentially* generate adversarial perturbations for RNNs. In other words, the $k^{th}$ input $\tilde{x}(k)$ of the *adversarial sequence* is generated in constant time using just the $k^{th}$ input $x(k)$ of the nominal sequence, and the $k^{th}$ hidden states. This means that sequential inputs can be adversarially perturbed one timestep at a time, rather than collecting the entire sequence. Figure 1 illustrates this point. This in principle, means that adversarial noise can be injected into real-time signals like speech commands. Our method is also amenable for scenarios with very long input sequences, as it would *scale linearly* with sequence length.
- We present a control theory-based analysis to explain how the proposed disturbance sequence, viewed as a feedback control law, steers the state trajectories of the adversarially perturbed system away from the nominal, unperturbed states. Also, viewed in this manner as a dynamical system, we show how the problem of finding adversarial perturbations can be formulated as an optimal control problem, which enables us to use an off-the-shelf control toolbox for crafting adversarial examples.
- Finally, a constructive proof regarding the existence of adversarial perturbations is provided. In particular, we present conditions under which one can always find an adversarial perturbation to a given nominal input. Since these results are analytical in nature, the conditions may be used in robust training of RNNs by directly imposing constraints on the training parameters.

In order to better elucidate the dynamical system analysis in the sections to follow, we begin with a concrete example of a sequence classification task, and how the RNN classifier can be seen as a time-varying nonlinear system, with disturbance acting as a control input. The formal analysis results will be presented in Section IV. Later, we shall experimentally demonstrate our approach of crafting adversarial perturbations to a variety of RNN classification examples covering different network architectures and sizes, in Section V, namely: (a) Frequency discrimination using Gated Recurrent Unit (GRU) architecture, (b) MNIST digit classification using vanilla RNN, (c) Human activity recognition using GRU, and (d) IMDb movie review classification using Long Short-Term Memory (LSTM) architecture.

## II. Time series classification task

One of the key applications of RNNs is sequence classification where the input-output map is "many-to-one." Commonly used examples in the literature include sentiment analysis (IMDB reviews or Twitter), MNIST digit recognition, human activity recognition (HAR), handwriting recognition (IAM On-Line Handwriting Database), urban sound classification etc. The example we use to motivate the discussion to follow, is the 'frequency discrimination task' [17], [18]. In this task, the RNN classifier is to classify sinusoidal input sequences based on their time period $T$ into two classes, as shown in Figure 2. Details on the model and training can be found in Section V, but in this section, we shall refer to this toy example to build a basic background.

## III. Perturbing the inputs for misclassification

The goal of an adversarial attack is to modify the inputs in a way that is indistinguishable to a human, but lead the neural

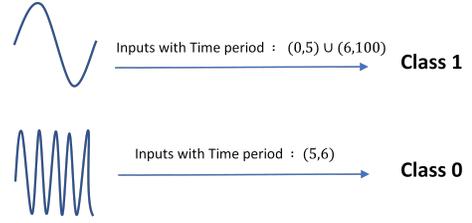

Fig. 2. Toy example of RNN classification task (Frequency discrimination).

network to have a very different behaviour. Let us start with the compact representation of a RNN:

$$h(k+1) = \hat{f}(h(k), x(k)), \quad h(0) = \text{fixed} \quad (1)$$
$$y(k) = \hat{g}(h(k)). \quad (2)$$

where $h(k) \in \mathbb{R}^n$ is the hidden state vector and $x(k) \in \mathbb{R}^m$ is the RNN input. The output vector $y(k) \in \mathbb{R}^l$ for an RNN classifier is interpreted as a probability distribution over the $l$ classes.
For the classification task in Figure 2, for example, the function $\hat{f}$ represents a GRU layer and $\hat{g}$ represents a softmax layer, with $n = 2, m = 1$, and $l = 2$. The phase plot of the network ($h(k)$ trajectories) corresponding to input signals belonging to the two classes are shown in Figure 3(a). Plotted in Figure 3(b) are the respective classification probabilities of signals (i.e., the first component of output vector $y(k)$ for class 0 signals and second component for class 1 signals), which appear to be uniformly lower bounded by some monotonically increasing function.

We convert our RNN equation to a continuous time system for the convenience of analysis as

$$\frac{dh}{dt} = f(h, x) = \frac{1}{\Delta}(\hat{f}(h, x) - h), \quad h(0) = h_0, \quad (3)$$

where the state $h(t) \in \mathbb{R}^n$, and the control input $x(t) \in \mathbb{R}^m$, and $\Delta$ is a small positive constant. Let us say the RNN is trained to classify input sequences belonging to set $\mathcal{U}$ into two classes. Further, let there be two sets of input signals, denoted by $\mathcal{U}_a, \mathcal{U}_b$ s.t. $\mathcal{U}_a \bigcup \mathcal{U}_b = \mathcal{U}$, and open sets $\mathbb{S}_a, \mathbb{S}_b \in \mathbb{R}^n$ such that if $x_a(t) \in \mathcal{U}_a$ then the corresponding state trajectory $h_a(t)$ of system (3) converges to the interior of the set $\mathbb{S}_a$. Similarly $h_b(t)$ corresponding to any $x_b(t) \in \mathcal{U}_b$ converges to $\mathbb{S}_b$. In our frequency discriminator example, the sets $\mathbb{S}_a, \mathbb{S}_b$ can be taken to be the two half-planes whose boundaries are the '0.99' and '0.11' lines respectively (shaded regions in green and blue), in Figure 3(a). Additionally, motivated by Figure 3(b), let us define a Lyapunov-like function $V(h)$ supported on $\mathbb{S}'_a$ such that $W(t) > V(h_a(t)) > 0$ whenever $h_a(t) \notin \mathbb{S}_a$, where $W(t)$ is a monotonically decreasing function.

Given input $x_a(t) \in \mathcal{U}_a$, our goal is to find a perturbation $d(t)$ to $x_a(t)$ such that the corresponding trajectory no longer converges to $\mathbb{S}_a$, and possibly, converges to $\mathbb{S}_b$. In other words, if we consider solution $h(t)$ given by

$$\dot{h}(t) = f(h(t), x_a(t) + d(t)), \quad h(0) = h_0 \quad (4)$$

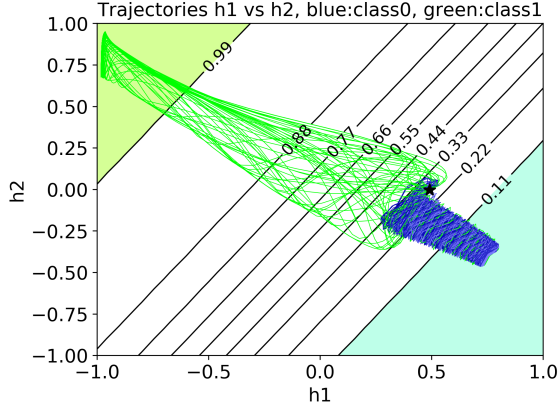

(a) Phase plot of RNN classifier for frequency discrimination example. $h1$ and $h2$ are the components of the two dimensional state vector $h$.

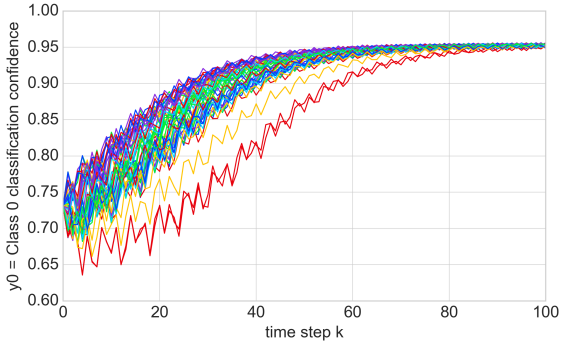

(b) RNN output vector $y = (y0, y1)$ denotes the classification probability corresponding to each of the two classes.

Fig. 3. (a) All the 100 trajectories (50 from each input class) start at the point $[0.4917, -0.0038]$ marked by the black star. Solid black lines show the level sets of the classification probability corresponding to class 1 (for example, the inputs corresponding to the green trajectories crossing the 0.99 line are classified as class 1 with more than 99% confidence). (b) Corresponding classifier confidence in classifying the class 0 inputs as class 0.

then we want $V(h(t)) > \epsilon$ for all times $t$, and some positive constant $\epsilon$. In order to lead the network to misclassification, we need a stronger condition, i.e., given a $x_a(t) \in \mathcal{U}_a$ we want $d(t)$ subject to $x_a(t) + d(t) \in \mathcal{U}_b$.

We start with

$$\begin{aligned}
\dot{h}_1(t) &= f(h_1(t), x_a(t)) \\
\dot{h}_2(t) &= f(h_2(t), x_a(t) + d(t)) \\
h_1(0) &= h_2(0)
\end{aligned} \quad (5)$$

along with

$$0 < V(h_1(t)) < W(t).$$

Then, along the trajectory $h_2(t)$, we have

$$\begin{aligned}
\dot{V}(h_2(t)) &= \left(\nabla_h V(h_2)\right)^T \dot{h}_2(t) \\
&= \left(\nabla_h V(h_2)\right)^T f(h_2, x_a + d) \\
&= \left(\nabla_h V(h_2)\right)^T \big(f(h_2, x_a) + \\
&\quad (\nabla_x f(h_2, x_a))d + \mathbf{o}(\|d\|)\big) \\
&\simeq \left(\nabla_h V(h_2)\right)^T f(h_2, x_a) +
\end{aligned}$$

$$\left(\nabla_h V(h_2)\right)^T (\nabla_x f(h_2, x_a))d. \quad (6)$$

Let us define the term $g(h, u) \doteq \left(\nabla_h V(h)\right)^T f(h, u)$, so the first term in the last line of equation (6) is equal to $g(h_2, x_a)$. Now, using the Mean-Value Theorem, we get

$$g(h_2, x_a) = g(h_1, x_a) + \left(\nabla_h g(\tilde{h}, x_a)\right)^T (h_2 - h_1),$$

for some $\tilde{h} = h_1 + \alpha(h_2 - h_1), \alpha \in (0, 1)$. Substituting this into equation (6) yields,

$$\begin{aligned}
\dot{V}(h_2(t)) &\simeq \left(\nabla_h V(h_2)\right)^T f(h_1, x_a) + \quad (7) \\
&\quad \left[\nabla_h^2 V(\tilde{h}) \cdot f(\tilde{h}, x_a) + \nabla_h f(\tilde{h}, x_a) \cdot \nabla_h V(\tilde{h})\right]^T (h_2 - h_1) \\
&\quad + \left(\nabla_h V(h_2)\right)^T (\nabla_x f(h_2, x_a))d \\
&= \underbrace{\dot{V}(h_1(t))}_{\leq 0} + \left(\nabla_h V(h_2)\right)^T (\nabla_x f(h_2, x_a))d + \\
&\quad \underbrace{\left[\nabla_h^2 V(\tilde{h}) \cdot f(\tilde{h}, x_a) + \nabla_h f(\tilde{h}, x_a) \cdot \nabla_h V(\tilde{h})\right]^T (h_2 - h_1)}_{\text{bounded}}.
\end{aligned}$$

At this point, we note that choosing

$$d = d(t) = \alpha(\text{sign}(\left(\nabla_h V(h_2)\right)^T (\nabla_x f(h_2, x_a))))^T, \quad (8)$$

for some small positive $\alpha$, would be an appropriate choice in making $V(h_2(t))$ increase (instantaneously) at time $t$. Indeed, we can make our frequency discriminator RNN misclassify and assign very high probability to the incorrect class by choosing such a disturbance, as shown in the next two figures.

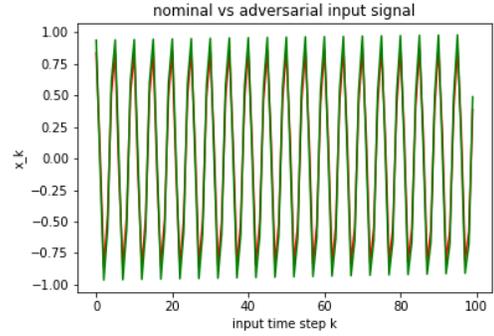

(a)

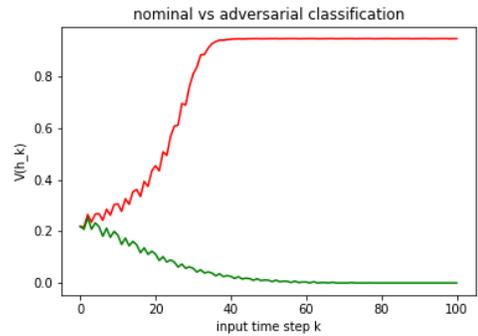

(b)

Fig. 4. (a) Input sequence belonging to class 0 in green (correctly classified with $> 95\%$ confidence) is misclassified as class 1 when perturbed slightly, in red (with very high confidence of $> 99\%$). (b) Lyapunov-like function decreases along the nominal trajectories (green), but increases along the perturbed one.

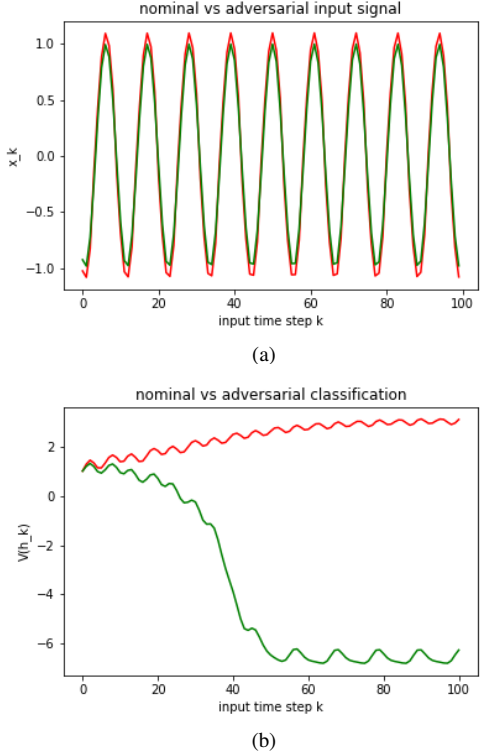

Fig. 5. (a) Input sequence belonging to class 1 in green (correctly classified with > 99% confidence) is misclassified as class 0 when perturbed slightly, in red (with very high confidence of > 95%). (b) Lyapunov-like function decreases along the nominal trajectories (green), but increases along the perturbed one.

Figures 4 and 5 show how the addition of the adversarial disturbance given by equation (8) (with $\alpha = 0.15$) to the nominal input impacts the network. The Lyapunov-like function $V$ shown in Figure 4 is chosen (among many other ways) as follows:

$$V(h) = \max\left(0, \begin{bmatrix} 1 \\ 0 \end{bmatrix}^T y(h) - \bar{y}\right)^2, \quad (9)$$

$\bar{y} = $ threshold $= 0.9$.

Since only the direction of $\nabla_h V$ is relevant, we can use any affine transformation of $V(h)$ given in equation (9), as indicated in Figure 5.

Before we move to the main result and continue with the analysis in the next section, it is important to note at this point that such a gradient-based attack on RNNs is different from previous work on adversarial attacks on feedforward NNs, like CNNs for image-classification. This is because the adversarial input disturbances are applied sequentially at each time, and the disturbance at time $k$ would depend on the disturbance at previous times implicitly through its dependence on the state trajectory $h(k)$ (as they are influenced by past input disturbances). For attacks in CNNs, for example, the disturbance to each pixel is computed simply based on a gradient around the current value of the pixels, and therefore these pixel-wise disturbances do not affect one another. Another practical difference in our disturbance computation for RNNs compared to previous methods, is that disturbances at time step $k$ are computed without any knowledge of future inputs. Previous attacks are limited in the sense that one would require to collate the inputs at all timesteps before computing the disturbance, thus prohibiting 'real-time' injection of disturbance to the inputs.

In the next section, we show the worst case effect of the input disturbance given by equation (8). The main result of this paper is also presented.

## IV. EXISTENCE OF ADVERSARIAL DISTURBANCE

As we mentioned earlier, in this section, we present the existence results for adversarial perturbations in RNNs. But let us first look at how the effect of disturbance (8) can be upper bounded. Robustification against adversarial attacks can then be translated to making this upper bound smaller.

Thus, returning back to equation (7), let us simplify the expressions by considering $V(h) = a^T h + b$, i.e., to be linear, so we get $\nabla V = a$ and $\nabla^2 V = 0$. One can then integrate both sides of equation (7) and obtain:

$$a^T h_2(t) = a^T h_1(t) \quad (10)$$
$$+ \int_0^t (\nabla_h f(\tilde{h}, x_a) a)^T (h_2(s) - h_1(s)) ds$$
$$+ \int_0^t \alpha \|a^T (\nabla_x f(h_2, x_a))\|_1 ds.$$

Using the Bellman-Grönwall inequality we then obtain:

$$|a|_{min} \cdot \|h_2(t) - h_1(t)\| \quad (11)$$
$$\leq \|a\| \int_0^t \underbrace{\|\nabla_h f(\tilde{h}, x_a)\|}_{\beta(t), \text{ bounded}} \|(h_2(s) - h_1(s))\| ds$$
$$+ \|a\| \underbrace{\int_0^t \alpha \|(\nabla_x f(h_2, x_a))\|_1 ds}_{\doteq \gamma(t)}$$
$$\implies \frac{|a|_{min}}{\|a\|} \|h_2(t) - h_1(t)\| \leq \gamma(t) \exp\left(\int_0^t \beta(s) ds\right),$$

where $|a|_{min}$ is the minimum over the absolute values of the elements of vector $a$. It may be of our interest to note that the term $\beta(t)$ which enters exponentially is more critical than the term $\gamma(t)$ which impacts the separation of the two trajectories only linearly, as is the case in linear systems. In this sense, the weights associated with the states more critically impact the state trajectories compared to the weights associated with the input.

### A. Lower-bounding trajectory separations using Matrix measures

Previously, we presented how a gradient based disturbance may reasonably lead the perturbed system to diverge from the nominal system. The upper bound in equation (11), defines the largest divergence that the input disturbance can produce, but since such estimates are norm based, they are usually conservative. Matrix measures provide a tighter estimation of system trajectories [19], since they are not norm based and can therefore be positive or negative.

The matrix measure of a matrix $A$, with induced norm $\|\cdot\|$ is defined as $\mu(A) \doteq \lim_{\theta \to 0^+} \frac{\|I+\theta A\| - \|I\|}{\theta}$, and always exists. Given a system

$$\dot{h} = A(t)h(t),$$

one may bound its solutions using Coppel's inequality as

$$\|h(t)\| \geq \|h(t_0)\| \exp \int_{t_0}^{t} -\mu(-A(s))ds \quad (12)$$

$$\|h(t)\| \leq \|h(t_0)\| \exp \int_{t_0}^{t} \mu(A(s))ds.$$

In order to be able to apply this to our system, we first start by writing equation (5) in terms of $e(t) \doteq h_2(t) - h_1(t)$ as

$$\begin{aligned}
\dot{e}(t) &= \dot{h}_2(t) - \dot{h}_1(t) = f(h_2, x+\delta) - f(h_1, x) \quad (13) \\
&= f(h_2, x) - f(h_1, x) + f(h_2, x+\delta) - f(h_2, x) \\
&= \left[\int_0^1 \nabla_h f(h_1 + \lambda e, x) d\lambda\right] e(t) \\
&\quad + \left[\int_0^1 \nabla_x f(h_1 + e, x + \lambda \delta) d\lambda\right] \delta(t) \\
&= A(t,e)e(t) + B(t,e,\delta)\delta(t),
\end{aligned}$$

and $e(0) = 0$. Once $e$ becomes non-zero, a sufficient conditions for the trajectories $h_1, h_2$ to continue diverging even without disturbance (i.e., $\delta$ set to zero), is when the matrix measure $\mu(-A(t,e))$ is negative for all $t,e$. This sufficient condition is easy to obtain in analytical form, as illustrated by the following example.

**Example 1.** Let us consider the vanilla RNN architecture given by

$$h(k+1) = \tanh(Uh(k) + Wx(k) + b)$$

which is approximated in continuous time as

$$\dot{h} = \frac{1}{T}(\tanh(Uh + Wx + b) - h).$$

Then, the term $A(t,e)$ is given by

$$\frac{1}{T}\int_0^1 \left(diag\left[1 - \tanh^2((1-\lambda)Uh_1 + \lambda Uh_2 + Wx + b)\right]U - I\right)d\lambda.$$

We now choose our vector norm to be the infinity norm, and define the matrix norm and the matrix measure using this norm. Then, we get

$$0 > \mu(-A(t,e)) \quad (14)$$
$$\Longleftarrow 0 > \mu\left(I - diag\left[1 - \tanh^2(Uh + Wx + b)\right]U\right)$$
$$\forall h \in [-1, 1]^n$$
$$\Longleftarrow 0 > \mu\left(diag\left[1 - \tanh^2(Uh + Wx + b)\right]^{-1} - U\right)$$
$$\forall h \in [-1, 1]^n.$$

We then obtain, $\left[1 - \tanh^2(U_i h + W_i x + b_i)\right]^{-1} + \sum_{j \neq i} |U_{ij}| < U_{ii}, \quad \forall h \in [-1,1]^n, \quad \forall i = 1, 2, ..., n \implies \mu(-A(t,e)) <$ 0. The inequality on the left-hand side can be numerically verified to hold whenever

$$2\|U_i\|_1 + W_i x + b_i + 0.8 < 2U_{ii}.$$

*B. Constructing the adversarial perturbation*

We now look at the disturbance term in last line of equation (13). Since the input disturbance appears in a nonlinear fashion on the left hand side of equation (13), we present the following proposition, which allows us to transform the input $\delta(t)$ into another input that enters the systems in an affine fashion.

*Proposition 1:* Consider the term $B(t,e,\delta)\delta$ on the left-hand side of equation (13). For any given bounded vector $v \in \mathbb{R}^n$ and $t \in \mathbb{R}$, there exist $w$ and some scalar $\alpha > 0$ such that

$$w = \alpha B(t,e,w)^T v$$

if $\nabla_x^2 f_i$ is bounded for every $i = 1, 2, ..., n$.

**Proof** Our proof relies on using the Contraction Mapping Theorem to show that the function $\mathcal{B}(z) = \alpha B(t,e,z)^T v$ defined on $\mathbb{R}^n$ has a (unique) fixed-point $w$, for some scalar $\alpha > 0$. If $\|\mathcal{B}'(z)\| = \alpha\|\frac{dB(t,e,z)^T v}{dz}\| < 1$ uniformly for all $z$, then the map $\mathcal{B}$ is a contraction. We then only need to show that $\|\frac{d}{dz}B(t,e,z)^T v\|$ is uniformly bounded, as $\alpha > 0$ can be chosen arbitrarily small to make the product less than 1. Clearly, $\frac{d}{dz}B(t,e,z)^T v = \sum_{i=1}^n v_i \int_0^1 \nabla_u^2 f_i(h_1 + e, x + \lambda z)\lambda d\lambda$. This means

$$\|\frac{d}{dz}B(t,e,z)^T v\| \leq \sum_{i=1}^n |v_i| \int_0^1 \|\nabla_x^2 f_i(h_1 + e, x + \lambda z)\|d\lambda$$

which is bounded when $v$ and $\nabla_x^2 f_i$ are bounded. $\square$

We can now use Proposition 1 to craft an adversarial perturbation as follows. For our system (5) (and any other RNN system in general), since the function $f$ satisfies the condition in Proposition 1 (on boundedness of the Hessian), then for any bounded $v(t) \in \mathbb{R}^m$ there exists a fixed-point $\bar{\delta}(t)$ of the function $\alpha B(t,e,\cdot)^T v(t)$ for some $\alpha > 0$. This fixed point may be computed as the limit of the system

$$\delta_{n+1} = \alpha B(t,e,\delta_n)^T v(t), \quad (15)$$

as $n \to \infty$ due to the Contraction Mapping Theorem. In particular, we can choose $v(t)$ to be equal to $e(t)$, which is bounded from above due to the fact that the states of system (5) evolve in a bounded region. Thus, we have

$$\bar{\delta}(t) = \alpha B(t,e,\bar{\delta})^T e(t) \quad (16)$$

and plugging this $\bar{\delta}$ as the disturbance in equation (13) gives us

$$\begin{aligned}
\dot{e}(t) &= A(t,e)e(t) + B(t,e,\bar{\delta})\bar{\delta}(t) \quad (17) \\
&= A(t,e)e(t) + \alpha B(t,e,\bar{\delta})B(t,e,\bar{\delta})^T e(t) \\
&= \left[A(t,e) + \alpha B(t,e,\bar{\delta})B(t,e,\bar{\delta})^T\right]e(t) \\
&\doteq K(t,e)e(t).
\end{aligned}$$

Clearly, the second term in (17) with the adversarial disturbance $\bar{\delta}$ leads $e(t)$ to diverge. This is because when we exclude the first term, we get

$$\dot{e}(t) = \alpha B(t, e, \bar{\delta}) B(t, e, \bar{\delta})^T e(t)$$
$$\implies e(t)^T \dot{e}(t) = \alpha \|B(t, e, \bar{\delta})^T e(t)\|^2$$
$$\implies \frac{d}{dt}\|e(t)\|^2 = 2\alpha \|B(t, e, \bar{\delta})^T e(t)\|^2 \geq 0.$$

We can now summarize the discussions in this subsection into the following theorem:

*Theorem 1:* Let us consider an RNN represented by the equation $\frac{d}{dt}h(t) = f(h(t), x(t))$, with some fixed initial state $h(0)$. For any given nominal input $x_1(t)$, and corresponding state trajectory $h_1(t)$, there exists some input perturbation $\delta(t)$ such that the state $h_2(t)$ of the perturbed system corresponding to input $x_1(t) + \delta(t)$ monotonically diverges if the matrix measure $\mu(-A(t, e)) < 0$.

Finally let us draw a connection between the "fixed-point" adversarial disturbance proposed in this section and the gradient based disturbance (8) described in Section III. It can be shown that one is the limiting case of the other. This to a certain extent helps experimentally corroborate the proofs and results in Proposition 1 and Theorem 1, since gradient based perturbation in the previous section was already shown to be an effective adversarial attack.

*Proposition 2:* The perturbation $\bar{\delta}$, computed using equation (16), approaches the gradient based perturbation

$$\delta(t) = (\nabla_x f(h_2, x_a))^T e(t)$$

in direction, (i.e., $\bar{\delta} \parallel \delta$). Thus, taking $V = \frac{1}{2}(h_2 - h_1)^2$, disturbance (8) and disturbance (16) have the same sign.

**Proof** We start by looking at the direction of $\bar{\delta}$ satisfying equation (16) as $\alpha$ approaches zero. Clearly, since $B(t, e, \cdot)^T e(t)$ is a bounded function, $\alpha \to 0 \implies \bar{\delta} \to 0$ which by continuity of $B$, implies $B(t, e, \bar{\delta})^T e(t) \to B(t, e, 0)^T e(t)$, and this further implies $\bar{\delta} \to \alpha B(t, e, 0) e(t)$.
Next we note that by definition, $B(t, e, 0) = \int_0^1 \nabla_x f(h_2, x) d\lambda = \nabla_x f(h_2, x)$. Thus, we get $\alpha \to 0 \implies \bar{\delta} \to \alpha \nabla_x f(h_2, x)^T e(t)$. This means $\bar{\delta} \parallel \nabla_x f(h_2, x)^T e(t)$ as $\alpha \to 0$. □

The next section presents a way to simplify the computation of the fixed-point disturbance and presents some analysis on how this new disturbance converges to the manifold given by (16), while leading $\|e\|$ to grow.

### C. Dynamic computation of adversarial disturbance

The dynamics of error $e(t)$ driven by the input disturbance $d(t)$ is described by a continuous-time differential equation (13), whereas the input disturbance (16) needs to be computed at each (continuous) time instant $t$ by solving the implicit algebraic equation (16). Although this computation can be done in a straightforward manner using the globally converging discrete-time difference equation (15) if performed at a timescale faster than the error dynamics, this discrepancy between continuous versus discrete dynamics needs to be handled in a more formal manner.

Our aim in this section is to derive an ordinary differential equation (ODE) $\frac{d\delta}{dt} = \phi(t, e, \delta)$ that determines the evolution of adversarial disturbance in continuous time, consistent with the continuous error dynamics. We would ideally want the solution of this ODE to evolve in the manifold given by equation (16). One way to achieve this, is perhaps with sliding mode controller $u = \phi(t, e, \delta)$ that takes $\delta(t)$ to the manifold (16) in finite time and keeps it on that manifold thereafter. However, such a controller would not be very encouraging as far as practical implementation is concerned, since we would need to compute gradients of each element of matrix $B$.

We start with the differential equation (13), along with the ODE

$$\epsilon \frac{d\delta}{dt} = \alpha B(t, e, \delta)^T e - \delta, \tag{18}$$

where $\epsilon$ is a positive constant. Let $\delta(t) = h(t, e)$ be the explicit solution of the equation (16), and $\delta(t, \epsilon)$ be the solution of equation (18). Note that by setting $\epsilon = 0$, one obtains what is called in control theory literature, a 'singularly perturbed' system [20] described by the original set of equations (13) and (16). So, it is reasonable to expect that the disturbance $\delta(t, \epsilon)$ would also act as an adversarial disturbance for sufficiently small values of $\epsilon$. We can now introduce a new variable $y(t, \epsilon) = \delta(t, \epsilon) - h(t, e)$. Then, the following holds.

*Theorem 2:* Given the systems (13) and (18)

$$\dot{e}(t) = A(t, e)e(t) + B(t, e, \delta)\delta(t)$$
$$\epsilon \dot{\delta}(t) = \alpha B(t, e, \delta)^T e - \delta,$$

the term $y(t, \epsilon)$ satisfies the exponentially decaying bound $\|y(t, \epsilon)\| \leq k_1 \exp\left(-\frac{k_2}{\epsilon}t\right)\|y_0\| + k_3 \epsilon \quad \forall t > 0$ globally i.e., for any initial condition $\|y_0\|$, and uniform constants $k_1, k_2, k_3 > 0$.
**Proof** Please see Appendix.

*Corollary 1:* Given any time $T > 0$, $\eta > 0$ and initial condition $y_0$, there exists a $\epsilon'$ such that $\|y(t, \epsilon')\| \leq \eta$ for all $t > T$.

Let us now look at how the disturbance $\delta(t, \epsilon)$ affects term $\|e(t)\|$. We recall that our goal is to find adversarial disturbances to the input such that the two trajectories $h_1$ and $h_2$ diverge. Focusing just on the disturbance term, we have

$$\begin{aligned}\frac{1}{2}\frac{d}{dt}\|e\|^2 &= e^T \dot{e} = e^T B(t, e, \delta(t, \epsilon))\delta(t, \epsilon) \\ &= e^T B(t, e, \delta(t, \epsilon))\big[\alpha B(t, e, \delta(t, \epsilon))^T e + \\ &\quad \delta(t, \epsilon) - \alpha B(t, e, \delta(t, \epsilon))^T e\big] \\ &= e^T B(t, e, \delta(t, \epsilon))\big[\alpha B(t, e, \delta(t, \epsilon))^T e + \\ &\quad y(t, \epsilon) + h(t, e) - \alpha B(t, e, y(t, \epsilon) + \\ &\quad h(t, e))^T e\big].\end{aligned}$$

From the definition of $h(t, e)$, we know that $h(t, e) = \alpha B(t, e, h(t, e))^T e$. Furthermore in the proof of Proposition 1,

$\alpha \|B(t, e, h(t, e))^T e\| \leq \frac{1}{2}$ by construction. Using these two facts, we get

$$\begin{aligned}
&\frac{1}{2}\frac{d}{dt}\|e\|^2 \\
&= e^T B(t, e, \delta(t, \epsilon))\big[\alpha B(t, e, \delta(t, \epsilon))^T e + y(t, \epsilon) \\
&\quad + h(t, e) - \alpha B(t, e, y(t, \epsilon) + h(t, e))^T e\big] \\
&= e^T B(t, e, \delta(t, \epsilon))\big[\alpha B(t, e, \delta(t, \epsilon))^T e + y(t, \epsilon) \\
&\quad + \alpha B(t, e, h(t, e))^T e - \alpha B(t, e, y(t, \epsilon) + h(t, e))^T e\big] \\
&\geq \alpha\|B(t, e, \delta(t, \epsilon))^T e\|^2 - \|B(t, e, \delta(t, \epsilon))^T e\| \cdot \|y(t, \epsilon)\| \\
&\quad - \|B(t, e, \delta(t, \epsilon))^T e\| \cdot \\
&\quad \|\alpha B(t, e, h(t, e))^T e - \alpha B(t, e, y(t, \epsilon) + h(t, e))^T e\| \\
&\geq \alpha\|B(t, e, \delta(t, \epsilon))^T e\|^2 - \|B(t, e, \delta(t, \epsilon))^T e\| \cdot \|y(t, \epsilon)\| \\
&\quad - \|B(t, e, \delta(t, \epsilon))^T e\| \cdot \left(\frac{1}{2}\|y(t, \epsilon)\|\right) \\
&= \|B(t, e, \delta(t, \epsilon))^T e\|\left(\alpha\|B(t, e, \delta(t, \epsilon))^T e\| - \frac{3}{2}\|y(t, \epsilon)\|\right).
\end{aligned}$$
(19)

We want the right-hand side of the inequality above to be non-negative and remain so after some point of time. From corollary 1, the term $\|y(t, \epsilon)\|$ can become arbitrarily small at an exponential decay rate, if we choose $\epsilon$ small enough. This decay of $y(t, \epsilon)$ is independent of $e(t)$. So if there exists some $\mu_1, \mu_2 > 0$ such that $\|e(t)\| > \mu_1 \implies \|\alpha B(t, e, \delta(t, \epsilon))\| > \mu_2$, then perhaps we can find an invariant set of the form $\left\{(t, e)\big|\|e\| > \mu_1, t > t'\right\}$ with $\|e\|$ increasing on this set. Indeed, such an argument can be made if the following mild condition is satisfied:

*Assumption 1:* $B(t, e, 0)^T e = f_x(h_1(t) + e, x(t))^T e$ is non-zero for all $t > 0$ and $e \neq 0$.

Then, we can assure that the right-hand side of equation (19) eventually becomes strictly positive because of the following. Say there exist a time $t' > 0$ and $\mu_1 > 0$ such that $\|e(t')\| > \mu_1$, (uniformly in $\epsilon$). Let us define $\mu_2 \doteq \min_{\|e\| > \mu_1, t > t'} \left\{\|\delta\| \big| \delta = B(t, e, \delta)^T e\right\}$. Note that by Assumption 1, this means $\mu_2 > 0$. Thus, we have $\alpha\|B(t', e(t'), h(t', e(t')))^T e(t')\| = \|h(t', e(t'))\| > \mu_2 > 0$. Next, by continuity, there exists $p, q > 0$ such that $\alpha\|B(t', e(t'), h(t', e(t')) + y')^T e(t')\| \geq p$ whenever $\|y'\| \leq q$. Using Corollary 1, we can always choose an $\epsilon$ small enough such that $\|y(t', \epsilon)\| < \min(q, \frac{2}{3}p)$, and therefore $\|\alpha B(t', e(t'), h(t', e(t')) + y(t', \epsilon))^T e(t')\| = \|\alpha B(t', e(t'), \delta(t', \epsilon))^T e(t')\| \geq p$ and therefore from the last line of equation (19), we get

$$\frac{1}{2}\frac{d}{dt}\|e\|^2 > \frac{p}{\alpha}(p - p) = 0,$$

for all $t > t'$. Thus, we concisely present the preceding analysis as the following theorem.

*Theorem 3:* If 1) Assumption 1 holds, i.e., $B(t, e, 0)^T e = 0$ only when $e = 0$, and 2) matrix measure $\mu(-A(t, e)) < 0$, then $\|e(t)\|$ diverges monotonically once it leaves the equilibrium point $e = 0$.

*Remarks on Assumption 1:* The condition that $B(t, e, 0)^T e = f_x(h_1(t) + e, x(t))^T e = 0$ only for $e = 0$, is very mild in the case when size of the states of the system is equal to the input size. In other words, when the Jacobian matrix $f_x$ is square. For example, in the case of the RNN architecture in Example 1, Assumption 1 can be easily verified to hold if the (square) weight matrix $W$ is full-rank.

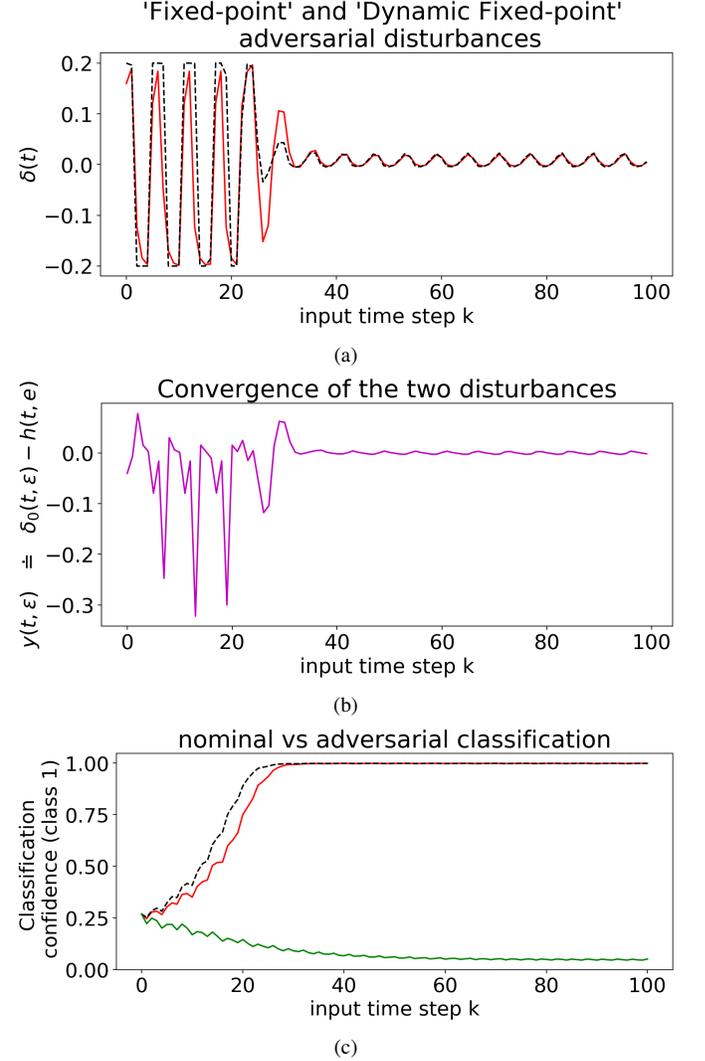

Fig. 6. (a) 'Fixed-point' disturbance (dashed black line) computed using equation (16) versus the dynamically computed disturbance (solid red line) using equation (18). (b) The solution of ODE (18) converges to the solution of the algebraic equation (16) as described in Theorem 2. (c) The green line shows that the class 0 nominal signal is correctly assigned a low confidence for class 1 ($\approx 0.05$) but assigned to class 1 with high confidence ($> 0.99$) after adding the input disturbances (dashed-black line ='Fixed-point disturbance' and solid-red line='Dynamic fixed-point disturbance').

Figure 6 illustrates an example, using the frequency discrimination task. A nominal input from class 0 injected with disturbances computed using equations (16) and (18), causing the network to misclassify it as class 1. In terms of runtime, the 'dynamic fixed-point attack' ((18)) was over 4× faster than the 'fixed-point attack' ((16)) in our Python implementation. The initial condition for ODE (18) was chosen to be zero, and the corresponding solution is denoted by $\delta_0(t, \epsilon)$ in the figures. One can verify that the bound on $\|y(t, \epsilon)\|$, described in

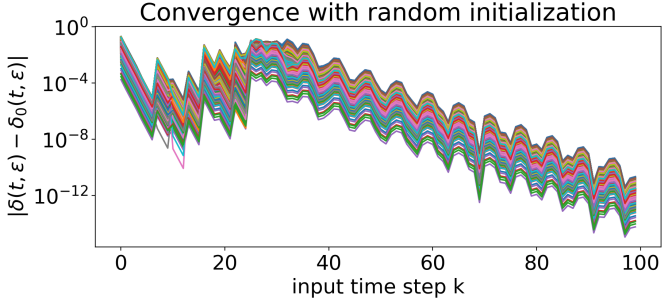

Fig. 7. The solutions of ODE (18) are computed for 1000 randomly sampled initial condition from a uniform distribution $\mathcal{U}(-1, 1)$. All of these $\delta(t, \epsilon)$ were able to fool the network.

Theorem 2, indeed holds globally by comparing $\delta_0(t, \epsilon)$ to the solutions of ODE (18) starting at different initial conditions, as show in Figure 7.

### D. Optimal adversarial disturbance computation

We have already demonstrated in Figures 4 and 5 with our simple frequency discriminator example that the gradient-based disturbance (8) can adversarially perturb inputs from both classes. But it may be of interest to explore optimally computed disturbances to evaluate how the approaches proposed in the previous subsections compare. Computation-wise, our proposed approaches are clearly cheaper, and disturbances can be injected in "real-time" since they only require the states and nominal input at the current timestep instead of the entire nominal state trajectory and input sequence. But it isn't clear how conservative our proposed method is, in terms of the size of the disturbance needed to fool the classifier. We present here our formulation for computing disturbances optimally, and a demonstration on frequency discrimination task of Figure 2. We first remind ourselves that the adversarial attack problem for RNNs can be reframed as a control synthesis problem. Given a nominal input sequence $x_a(t)$ for which we desire to generate an adversarial input $d(t)$, we can consider the system (4) as a nonlinear, time-varying control system with disturbance $d(t)$ as the control input. With such a system, we can use some well-developed computational tools from optimal control theory to generate adversarial disturbances. More concretely, we can solve the following:

$$\min_{\delta(t)} -V(h(T)) + \int_0^T \delta(s)^T R(s) \delta(s) ds,$$

subject to :

$$\frac{d}{dt} h = \bar{f}(t, h, \delta) = f(h(t), x_a(t) + \delta(t)), \ h(0) = h_0.$$

In order to force the adversarial disturbance to be bounded within some prescribed $\epsilon > 0$, we reparameterize the perturbation term as $\epsilon \tanh(\delta(t))$ in place of $\delta(t)$.

Picking $R = 1$ and $\epsilon = 0.15$, we obtain these optimal perturbations for misclassifying a randomly generated input signal belonging to class 1 as class 0. The corresponding gradient-based perturbation given by equation (8) is also computed. Figure 8 shows results of the comparison. The sign change in the gradient-sign based perturbation appears to coincide with the optimal perturbation, and performs equally well in leading the network to misclassify. It is interesting to note in Figure 8(c) that the gradient-sign based perturbation is more greedy in driving the network to misclassify, as one might expect.

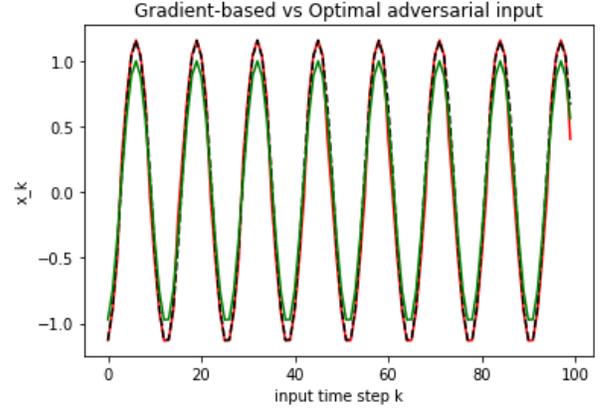

(a)

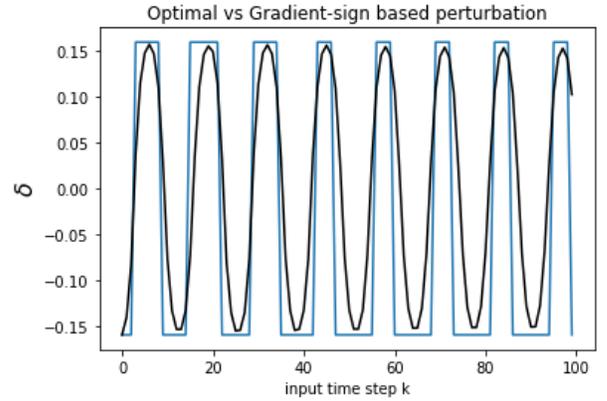

(b)

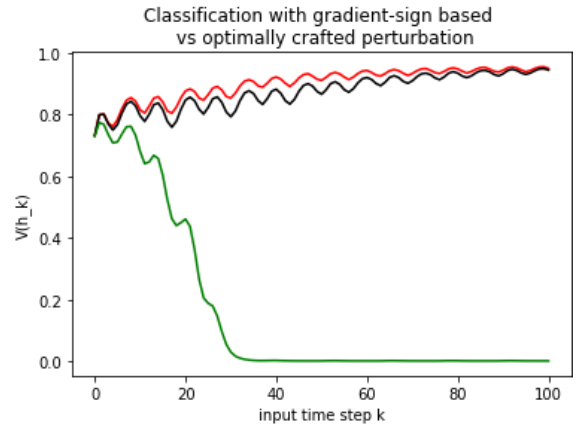

(c)

Fig. 8. (a) Nominal input sequence belonging to class 1 is shown green. The red and black dashed lines correspond to the gradient based and optimally perturbed inputs, respectively. (b) The perturbations added to the nominal input are shown in blue (gradient-sign) and black (optimal). (c) The nominal input is correctly classified with $> 99.8\%$ confidence (green), whereas the perturbed inputs are misclassified with very high confidence of $> 94\%$ (red = gradient-sign, black = optimal perturbation).

This concludes our presentation on control systems-rooted methodologies for constructing adversarial input disturbances. In the following section, we demonstrate these approaches on a few different applications involving RNN based classifiers.

TABLE I
SUMMARY OF THE THREE PROPOSED ADVERSARIAL DISTURBANCES

| Perturbation Method | $d(k+1)$ depends on | Compute time | Attack strength | $d(k)$ $l_2$-norm |
|---|---|---|---|---|
| Gradient based | $B(k+1, e(k+1), 0)$ | ★★★ | ★★★ | ★☆☆ |
| Fixed-point disturbance | $B(k+1, e(k+1), \delta(k+1))$ | ★☆☆ | ★★★ | ★★★ |
| Dynamic Fixed-Point | $B(k, e(k), \delta(k))$ | ★★☆ | ★★☆ | ★★★ |

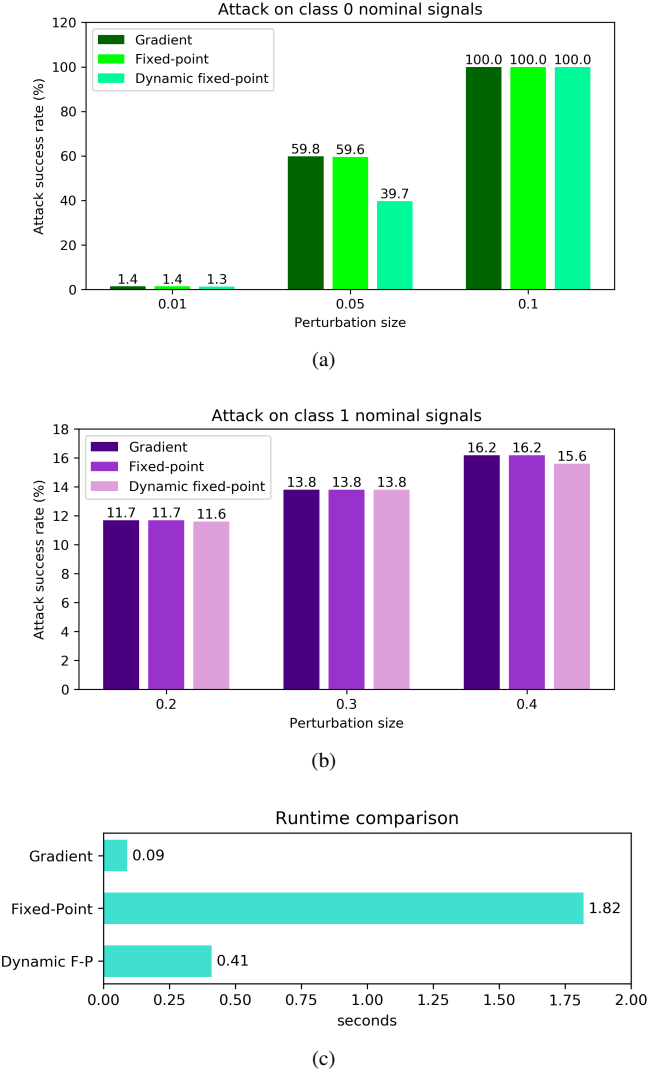

(a)

(b)

(c)

Fig. 9. Comparison between the three different adversarial disturbance computation methods

## V. EXPERIMENTS

### A. Frequency discrimination task

We now revisit the the frequency discrimination task [17], [18] that we briefly introduced in Section II. In that task, a sine wave is provided as an input to the RNN which then classifies it based on its time period. Specifically, sine inputs belong to one of the two classes: class 0 if the time period is in $(5, 6)$ or class 1 if the time period lies in $(0, 5) \cup (6, 100)$. Every sine input with period $T$ has a random phase shift drawn from $\mathcal{U}(0, T)$. The sequence length is 100, and the sinusoids are sampled at intervals of 0.1.

These inputs enter a GRU layer with hidden states with dimension 2 followed by a fully-connected softmax layer with output dimension 2, which represents the classification probabilities for the two classes. This network is trained using the standard cross-entropy loss, with equal number of generated training examples for each class, and has a test accuracy of $\approx 97\%$.

Throughout the previous sections, we have already seen in action various attacks (gradient-based, 'fixed-point', 'dynamic fixed-point', and optimal) on this model. Therefore, we shall instead compare our proposed methods here.

We test our attacks on 1000 randomly generated input signals from each class. Adversarially perturbing nominal input from class 0 seem to be much easier than perturbing class 1 inputs as indicated by Figure 9 (a) and (b), where a higher attack success rate is achieved with smaller disturbances for class 0 inputs. Figure 9 (c) compares the runtimes of the three methods. The gradient-based method is the fastest requiring only a single Jacobian computation at each timestep. The dynamic fixed-point method requires a single evaluation of the $B(t, e, \delta)$ matrix, which in turn requires multiple computations of the Jacobian matrix for numerical integration, and thus takes longer than the gradient-based attack. Finally, the fixed-point attack requires multiple evaluations of the $B(t, e, \delta)$ matrix in the fixed-point iteration given by equation (15) and thus is the slowest, as expected. A summary based on these results is presented in table I.

### B. MNIST digit classification

We use MNIST handwritten digit recognition as a second example here, where the goal is to add small disturbances to the pixels of a greyscale image of a digit between zero and nine so that the RNN classifier misclassifies it. Although RNNs are not a traditional choice of neural networks for this task, they can nevertheless be used for image classification tasks, by inputting the image one row of pixels at a time thus resembling a sequence of inputs.

The model consists of an input layer of size 28 representing rows of a 28x28 pixels image, followed by a vanilla RNN layer with hidden state dimension of size 56. This layer then connects to a dense softmax layer of output dimension 10 corresponding to the ten classes representing digits 0 to 9. The network is trained on 55k images, and achieves an accuracy of 96% on 10k test images.

Figure 10 shows how the addition of the perturbations to each pixel causes the network to misclassify the images. The magnitude of the perturbation added to each pixel is upper bounded by 0.07, and the attack is untargetted.

As we increase the magnitude of the pixel perturbations, we see that the network deteriorates in accuracy. Table II shows the classification accuracy on a set of 10k images obtained by adversarially perturbing the original test dataset.

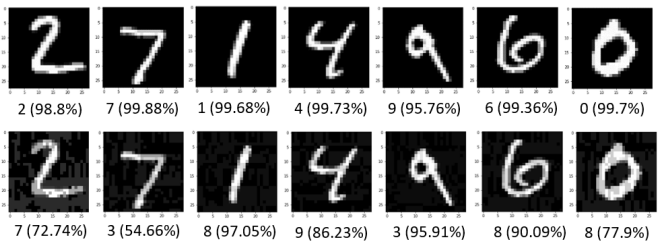

Fig. 10. Top row : Correctly classified, unperturbed digits. Bottom row: Perturbed images that were misclassified. The number indicates the class assigned to the image, and the percentage is the corresponding classification probability.

TABLE II
CLASSIFICATION ACCURACY ON ADVERSARIAL INPUTS

| Perturbation size | 0 | 0.05 | 0.1 | 0.2 | 0.3 |
|---|---|---|---|---|---|
| Accuracy | 96% | 61.34% | 29.76% | 12.83% | 10.29% |

### C. Human Activity Recognition

Human Activity Recognition (HAR) involves the identification of actions carried out by a person using observations of themselves and their environment. This typically involves inertial information from wearable sensors and the set of activities to be identified may include walking, sitting, laying down etc. In this experiment, we use the HAR dataset from the UCI Machine Learning Repository [21] to train a single-layer GRU network to recognize six different activities, namely - walking, walking upstairs, walking downstairs, standing, sitting, and laying down.

The input to the network is a 9-dimensional time-series data containing filtered sensor signals from tri-axial accelerometer and gyroscope. The network consists of a single GRU layer with hidden-state dimension of 50, followed by a dropout layer with a rate of 0.5. This is then followed by a dense ReLU-layer with output dimension of 50, and finally a dense softmax-layer with a 6-dimensional output representing a distribution over the six activity classes. The trained network achieves a test accuracy of 93.41%.

To illustrate the fixed-point based adversarial attack on this network, we consider a targeted attack wherein an input sequence corresponding to the activity 'standing' is perturbed such that the network classifies the signal as a different, pre-specified, target class ('walking'). Figure 11 shows the adversarial perturbation. The nominal input correctly classified as 'standing' (with probability $> 99.99\%$), is classified as 'walking' (with probability $> 99.99\%$) after injecting the adversarial perturbation.

Table III shows targeted attack success for 300 nominal input signals sampled randomly from the test dataset. We only consider those nominal signals for attack that the network correctly classified without any perturbations, and hence the diagonal entries are left out since they are by default 100%.

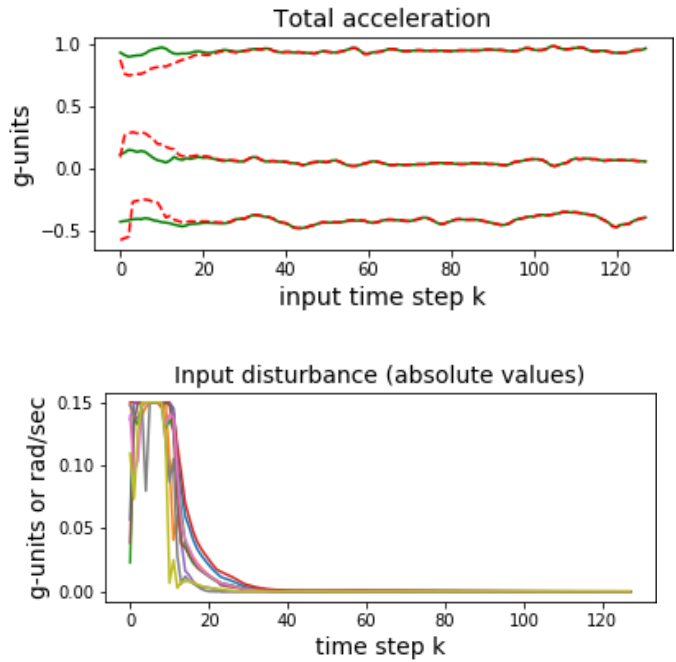

Fig. 11. Standing misclassified as Walking. **(Top)** Adversarially perturbed accelerometer signals (in dashed red) vs corresponding nominal, unperturbed signals (in solid green). **(Bottom)** Perturbations added to the 9-dimensional input signal (accelerometer + gyroscope) corresponding to 'standing'. The $l_\infty$ norm of the disturbance is constrained to lie below 0.15.

TABLE III
TARGETTED ATTACK SUCCESS RATE (%). INTEGER LABELS ARE AS FOLLOWS. 0: WALKING, 1: WALKING UPSTAIRS, 2: WALKING DOWNSTAIRS, 3: SITTING, 4: STANDING, 5: LAYING

| Target class → Nominal class ↓ | 0 | 1 | 2 | 3 | 4 | 5 |
|---|---|---|---|---|---|---|
| 0 | - | 54.84 | 43.55 | 17.74 | 20.97 | 0 |
| 1 | 46.15 | - | 30.77 | 26.92 | 3.85 | 5.77 |
| 2 | 50 | **86.11** | - | 5.56 | 0 | 0 |
| 3 | 11.36 | 11.36 | 6.81 | - | **97.73** | 0 |
| 4 | **77.27** | 22.73 | 29.55 | **100** | - | 0 |
| 5 | 0 | 12.9 | 0 | 16.13 | 0 | - |

The untargetted success rates for each of the classes 0 to 5 are respectively, 88.71%, 84.61%, 86.11%, 100%, 100%, and 20.97% (an untargetted attack is considered successful if the network misclassifies the nominal input to *any* one of the remaining classes). It is interesting to note the high success rates of targetted attacks 'standing' to 'sitting' and 'sitting' to 'standing', (even after considering error rates of 4.1% and 18.5%, respectively, for unperturbed inputs). This is consistent with the fact that sitting and standing are very close to one another, as physical activities.

### D. IMDb review sentiment analysis

Sentiment analysis aims at interpreting the subjective information such as sentiment expressed within textual input data, and classify them into classes such as positive, negative, or neutral, for example. In this example, we use the Stanford Large Movie Review Dataset [22] consisting of highly polar positive and negative movie reviews on IMDb. We use 25000 labelled data for training and testing our model, which we describe below.

Each textual input consists of a sequence of words and special characters which are converted into real-valued vectors of dimension 50, using the pre-trained GloVe word embedding, with 6 billion tokens and vocabulary size of 400k [23]. These embedding vectors are then inputted to a single-layer LSTM with state dimension of 128 (64 hidden and 64 memory states). A dropout level of 0.75 is applied to the output of this layer, which is followed by a final softmax layer with output dimension equal to 2. The test accuracy of the network is 82.7%.

Before we proceed, we would like to highlight an implementational challenge that separates this particular example from the previously demonstrated examples. Although the inputs to the LSTM are real-valued vectors, the raw input at each timestep belongs to a finite set of words, which means the word embedding vectors belong to a discrete and finite set $\mathcal{W} \subset \mathbb{R}^{50}$.

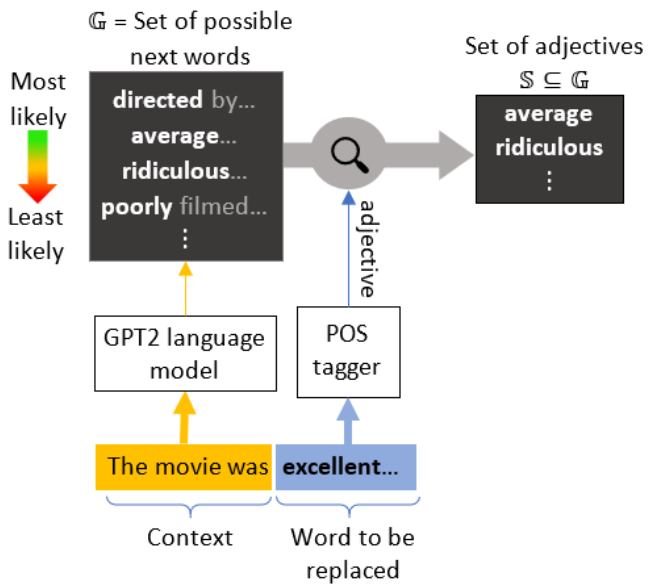

Fig. 12. Obtaining the set of 'replacement words' for each word in the original review.

Therefore, at any given timestep $k$, the computed adversarial input $\tilde{x}(k)$ corresponding to nominal input $x(k) \in \mathcal{W}$ to the LSTM layer may not belong to the set $\mathcal{W}$. To overcome this issue, we may use a heuristic similar to [11], wherein after computing $\tilde{x}(k)$, we pick the adversarial input to be equal to

$$\arg\min_{z \in \mathcal{V}}\{\|z - x(k)\|\},$$

where $\mathcal{V} = \{v \in \mathcal{W} | (v - x(k))^T(\tilde{x}(k) - x(k)) \geq 0\}$, which simply means that we pick an embedding vector closest to the computed adversarial input that aligns with the perturbation direction. A second more important challenge however, is to ensure the new adversarial input, when finally mapped back to a sentence, remains reasonably readable and fits well with the rest of the sentence. We achieve this by incorporating the state-of-the-art language model known as GPT-2 [24], as well as part-of-speech (POS) tagging provided within the NLTK package [25] to refine the set $\mathcal{V}$ so that newly generated adversarial sentence appears linguistically correct. Figure 12 explains how the set of possible replacement words at each timestep $k$ is generated. The set $\mathbb{S}$ consists of words that are similar to and contextually fitting replacements for the current word $k$, given a $\tau$ previous words $x(k-1), x(k-2), ..., x(k-\tau)$ in the original review. We choose $\tau$ to be equal to 4 for our experiments. Furthermore, since words in the set $\mathbb{S}$ are sorted by relevance, we can pick the top $M$ words in $\mathbb{S}$, denoted by $\mathbb{S}_M$. $M$ is chosen to be 20 in our experiment. Given the computed $\tilde{x}(k)$ obtained from equation (16), we then pick the replacement word to the original word $x(k)$ to be:

$$\arg\min_{z \in \mathcal{V}}\{\|z - x(k)\|\},$$

where the set $\mathcal{V}$ now is $\{v \in \mathbb{S}_M | (v - x(k))^T(\tilde{x}(k) - x(k)) \geq 0\}$. Note that for this particular application, since the set of admissible perturbations ($\mathbb{S}_M$) is a finite set, one may directly search of the best replacement word for $x(k)$ by looking at the effect that each of the words in $\mathbb{S}_M$ has on the current output $k$ of the classifier.

Tables IV and V illustrate adversarial text generation for sentiment classification. The original reviews are correctly classified with a very high confidence. The perturbations in this case are the replacement words, shown as orange colored text in the original review and blue colored text in the perturbed review. The perturbed review is then classified as negative, with very high confidence.

TABLE IV
EXAMPLE OF ATTACK ON NEGATIVE REVIEW

| | |
|---|---|
| Original review (−ve) classified negative (99.99%) | this film is really bad, so bad that even christopher lee cannot save it. a poor story an even poorer script and just plain bad direction makes this a truly outstanding horror film, the outstanding part being that it is the only horror film that i can honestly say i would never ever watch again. this garbage make plan nine from outerspace look like oscar material. |
| Perturbed review classified positive (97.88%) | this film is really fantastic, so forgive that even christopher lee cannot believe it. a poor story an utterly poorer script and slightly plain bad direction adds this a lovely innovative horror film, the next part being that it is the only horror film that i can honestly say i would never ever watch again. this garbage make plan nine from outerspace look like oscar material. |

Looking closer at the original and perturbed reviews in Table IV, the first occurrence of the word "bad" in the original negative review is replaced by "fantastic", however, the attack doesn't simply replace 'negative' sounding words with positive words as indicated the other two occurrences of the word "bad" in the original review. The perturbed review still retains negative sounding parts of the original review, and yet is able to fool the LSTM classifier.

TABLE V
EXAMPLE OF ATTACK ON POSITIVE REVIEW

| Original review (+ve) classified positive (99.67%) | this film is a great example of fine storytelling. the acting is superb. the story is inspiring without being overly manipulative or fake. there were a couple points where they probably made people a tad more good or bad than they really were, but considering it is a hollywood movie, they showed amazing restraint. there wasnt a single explosion shown in the movie, even though they had one opportunity to. the film, while having suspenseful parts, was not made into an action movie. the story is thus made to focus on an extraordinary man in unfortunately ordinary times. well done! |
|---|---|
| Perturbed review classified negative (99.94%) | this film is a massive example of fine storytelling. the acting is superb. the story is unfolding without losing directly manipulative or fake. politically reported a couple incidents where they already expected changes a tad better complex or negative than they initially suffered, but starting it is a hollywood movie, they showed playing restraint. really wasnt a serious explosion rocked in the movie, originally though they included 50 opportunity to. the film , while writing suspenseful lines, was already recovered into an action movie. the story is extremely flawed to focus on an extraordinary man in double terrible moments. well done! |

In these experiments, we have chosen to create our perturbations by replacing words in the original review by other, similar words that fit the context of the overall sentence. There still seems to be instances where our choice of replacement words does not lead to meaningful sentences, however. Other works in literature, such as [26], [27], have considered introducing perturbations into the reviews by incorrectly spelling out the original words. Since all the misspelled words get mapped via the word embedding to the same vector representing a 'unknown token' (denoted as 'unk'), the goal is then to find positions in the original review where we can place this 'unk' vector to adversarially impact the network. The original words in those positions are then simply replaced by their misspelled versions. Such perturbations can however be easily detected by any spell-checking software, but still are an interesting type of textual perturbation.

We would like to stress our algorithm iterates through the original review only once, and replaces words in the original review (if at all) *as it sees them in sequential order*, unlike the algorithm proposed in [11], which iterates through the entire sentence, changing one word per iteration until the review is misclassified.

## VI. CONCLUSION

Adversarial examples for neural network based classification have received continued interest among the learning community and yet, have lacked adequate analytical treatment. In this paper, we have provided sufficient conditions for existence of adversarial perturbations to any given input sequence to RNNs. Such perturbations can be constructed easily, and under a limiting condition are equivalent to a closed-form, gradient based perturbation. Furthermore since our formulation and analysis is inspired by control theory, our proposed adversarial perturbations can be constructed dynamically, at each time step, as the RNN gets its input sequentially, which is advantageous in two major ways. First, we do require to know the entire input sequence *a priori* for computing these perturbations. And second, our proposed method for crafting these adversarial additive perturbations to an input sequence scales linearly with the length of the sequence. Additionally, we take advantage of our dynamical systems based approach to show how optimal control may be used for computing adversarial perturbations. We illustrate this with some classification examples with varying complexities, in terms of architecture, and size of inputs and hidden states.

## APPENDIX

**Proof of Theorem 2:** We start by writing the dynamics of the variable $y(t, \epsilon) = \delta(t, \epsilon) - h(t, e)$ and $e(t)$ as follows:

$$\begin{aligned}
\frac{de}{dt} &= F(t, e, y) \doteq A(t, e)e(t) \\
&\quad + B(t, e, y + h(t, e))(y + h(t, e)), \\
\epsilon \frac{dy}{dt} &= G(t, e, y, \epsilon) \\
&\doteq \alpha B(t, e, y + h(t, e))^T e - y - h(t, e) - \epsilon \frac{\partial h}{\partial t}(t, e) \\
&\quad - \epsilon \frac{\partial h}{\partial e}(t, e) F(t, e, y).
\end{aligned}$$

Let us consider $t$ and $e(t)$ appearing in $G(t, e, \delta, \epsilon)$ to be 'frozen parameters' and look at the evolution of $y$ on a fast timescale $\tau = \epsilon t$:

$$\frac{dy}{d\tau} = G(t, e, y, 0). \qquad (20)$$

If we now take the Lyapunov function $V(y) = \frac{1}{2}\|y\|^2$, then along the solution $y(\tau)$ of equation (20), we have

$$\begin{aligned}
\dot{V} &= y(\tau)^T \frac{dy}{d\tau} = y^T G(t, e, y, 0) \\
&= -\|y\|^2 + y^T \left( \alpha B(t, e, y + h(t, e))^T e - h(t, e) \right) \\
&= -\|y\|^2 + y^T \left( \alpha B(t, e, y + h(t, e))^T e \right. \\
&\quad \left. - \alpha B(t, e, h(t, e))^T e \right) \\
&\leq -\|y\|^2 + \|y\| \cdot \|\alpha B(t, e, y + h(t, e))^T e \\
&\quad - \alpha B(t, e, h(t, e))^T e\| \\
&\leq -\|y\|^2 + \frac{1}{2}\|y\|^2 = -\frac{1}{2}\|y\|^2 = -V,
\end{aligned}$$

which implies that $y(\tau)$ is uniformly exponentially stable, and satisfies the inequality $\|y(\tau)\| \leq \|y(0)\| \exp\left(-\frac{1}{2}\tau\right)$. Using the (easily verifiable) fact that $\frac{\partial}{\partial y} G(t, e, y, 0)$ has bounded partial derivatives with respect to $t$ and $e$, along with the fact that

$G(t, e, 0, 0) = 0$ for all $t$ and $e$, we can obtain the following bounds:

$$\left\|\frac{\partial G}{\partial y}(t, e, y, \epsilon)\right\| = \left\|\frac{\partial G}{\partial y}(t, e, y, \epsilon) - \frac{\partial G}{\partial y}(t, e, 0, \epsilon)\right\|.$$

Thus, we can use Lemma 9.8 in [20] on slowly-varying systems, to conclude the existence of a Lyapunov function $V(t, e, y)$ such that

$$c_1 \|y\|^2 \leq V(t, e, y) \leq c_2 \|y\|^2, \quad (21)$$
$$\frac{\partial V}{\partial y} G(t, e, y, 0) \leq -c_3 \|y\|^2,$$
$$\left\|\frac{\partial V}{\partial y}\right\| \leq c_4 \|y\|, \quad \left\|\frac{\partial V}{\partial t}\right\| \leq c_5 \|y\|^2, \quad \left\|\frac{\partial V}{\partial e}\right\| \leq c_6 \|y\|^2.$$

Taking the time derivative of $V(t, e, y)$ we get,

$$\begin{aligned}\frac{dV}{dt}(t, e, y) &= \frac{\partial V}{\partial t} + \frac{\partial V}{\partial e} F(t, e, y) + \frac{1}{\epsilon}\frac{\partial V}{\partial y} G(t, e, y, \epsilon) \\ &\leq \frac{\partial V}{\partial t} + \frac{\partial V}{\partial e} F(t, e, y) + \frac{1}{\epsilon}\frac{\partial V}{\partial y} G(t, e, y, 0) \\ &\quad + \frac{1}{\epsilon}\frac{\partial V}{\partial y}\bigl(G(t, e, y, \epsilon) - G(t, e, y, 0)\bigr).\end{aligned}$$

Now, using the estimates $\|F(t, e, y, \epsilon)\| \leq k'$ and $\|G(t, e, y, \epsilon) - G(t, e, y, 0)\| \leq \epsilon L_3$, along with the norm bounds in equation (21), one obtains

$$\begin{aligned}\frac{dV}{dt}(t, e, y) &\leq c_5 \|y\|^2 + c_6 k' \|y\|^2 - \frac{c_3}{\epsilon}\|y\|^2 + L_3 c_4 \|y\| \\ &\leq -\frac{c_3}{2\epsilon}\|y\|^2 + L_3 c_4 \|y\|,\end{aligned}$$

if we choose $\epsilon \leq \frac{c_3}{2c_5 + 2c_6 k'}$. Thus, from the above inequality, one can bound $y(t, \epsilon)$ as follows:

$$\|y(t, \epsilon)\| \leq \sqrt{\frac{c_2}{c_1}} \exp\left(-\frac{c_3}{4c_2 \epsilon}t\right)\|y_0\| + \frac{2c_2 c_4 L_3}{c_1 c_3}\epsilon,$$

for all times $t > 0$ and initial condition $\|y_0\| \leq \sqrt{\frac{c_1}{c_2}}$. This completes the proof. □